\documentclass[final]{cvpr}

\usepackage{times}
\usepackage{epsfig}
\usepackage{graphicx}
\usepackage{amsmath}
\usepackage{amssymb}

\usepackage[pagebackref=true,breaklinks=true,colorlinks,bookmarks=false]{hyperref}

\usepackage{tabulary,multirow,xspace}
\usepackage{fixmath,mathtools,nicefrac,mmstyle}
\usepackage{cite}
\usepackage{booktabs}
\usepackage{subcaption}
\usepackage{caption}
\usepackage{multirow}
\usepackage{mmstyle}
\usepackage[misc]{ifsym} % for the envelop mark
\usepackage[title]{appendix} % for appendix
\newcommand{\tabincell}[2]{\begin{tabular}{@{}#1@{}}#2\end{tabular}}

\def\cvprPaperID{2940} % *** Enter the CVPR Paper ID here
\def\confName{ICCV}
\def\confYear{2021}

\begin{document}

%%%%%%%%% TITLE
\title{Exploring Data Augmentation for Multi-Modality 3D Object Detection}

\author{Wenwei Zhang$^1$ \quad Zhe Wang$^{2}$ \quad Chen Change Loy$^{1\textrm{\Letter}}$ \\
$^{1}$S-Lab, Nanyang Technological University \quad  $^{2}$SenseTime Research\\
%Institution1 address\\
{\tt\small $\left \{\text{wenwei001, ccloy} \right\}$@ntu.edu.sg \hspace{10pt} $\text{wangzhe}$@sensetime.com}}

\maketitle

%%%%%%%%%%%%%%%%%%%%%%%%%%%%%%%%%%%%%%%%%%%%%%%%%%%%%%%%%%%%%%%%%%%%%%%%%%%%%%%%%%%%%%%%%%%%%%%%%%%%
% !TEX root = ../main.tex
\begin{abstract}
% Three-dimensional (3D) object detection is essential in autonomous driving.
%
% There are observations that 
It is counter-intuitive that multi-modality methods based on point cloud and images perform only marginally better or sometimes worse than approaches that solely use point cloud.
This paper investigates the reason behind this phenomenon.
Due to the fact that multi-modality data augmentation must maintain consistency between point cloud and images, recent methods in this field typically use relatively insufficient data augmentation.
This shortage makes their performance under expectation.
Therefore, we contribute a pipeline, named transformation flow, to bridge the gap between single and multi-modality data augmentation with transformation reversing and replaying.
% to make any invertible single-modality augmentations applicable for multi-modality detection.
%
In addition, considering occlusions, a point in different modalities may be occupied by different objects, making augmentations such as cut and paste non-trivial for multi-modality detection.
We further present Multi-mOdality Cut and pAste (MoCa), which simultaneously considers occlusion and physical plausibility to maintain the multi-modality consistency.
% Considering the 
% which is more difficult considering the
% which selectively copy and paste patches 
%
% It boosts detection performance by cutting point cloud and imagery patches of ground-truth objects and pasting them into different scenes.
%
% in all the modalities during paste operation 
%
Without using ensemble of detectors, our multi-modality detector achieves new state-of-the-art performance on nuScenes dataset and competitive performance on KITTI 3D benchmark.
Our method also wins the best PKL award in the 3rd nuScenes detection challenge.
Code and models will be released at \url{https://github.com/open-mmlab/mmdetection3d}.
% \vspace{-0.5cm}
\end{abstract}

% !TEX root = ../main.tex
\section{Introduction}\label{sec:Introduction}

\begin{figure}[t]
	% \begin{minipage}{0.5\textwidth} 
	\centering\includegraphics[width=\linewidth]{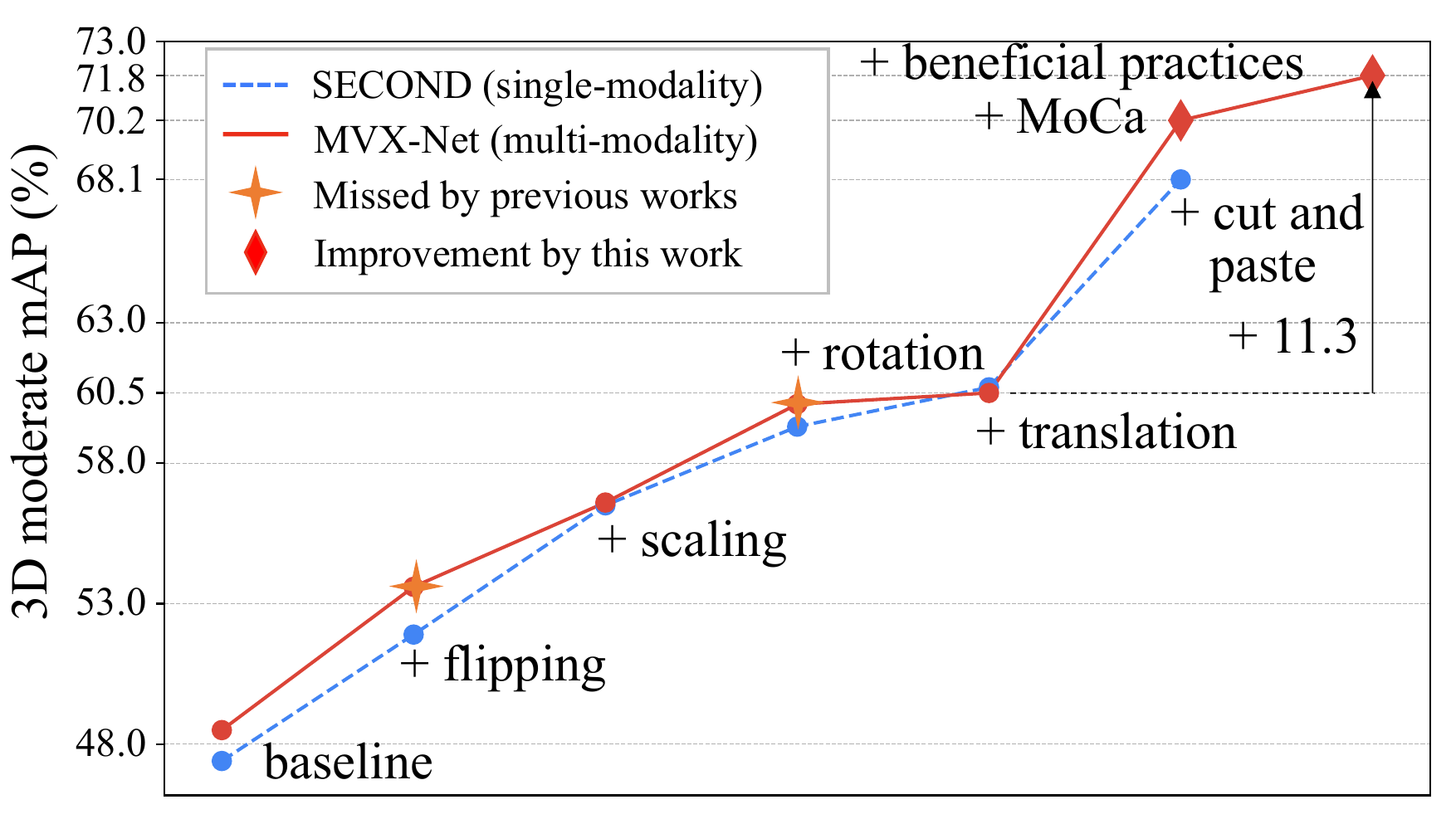}
	% \end{minipage}\hspace{0.01\textwidth}
	\vspace{-21pt}
	\caption{\small{\textbf{Step-by-step performance improvements brought by multi-modality augmentations, MoCa, and beneficial practices.}
			Random flipping, scaling, rotation, and translation enabled by the \emph{transformation flow} are equally effective for both SECOND and MVX-Net.
			\emph{Multi-modality cut and paste (MoCa)} with beneficial practices further improves the performance of MVX-Net,
			surpassing its single-modality counterpart (SECOND with cut and paste) by a large margin on KITTI dataset.
		}}\label{fig:aug_compare}
	\vspace{-18pt}
\end{figure}

Three-dimensional (3D) object detection is an essential vision task with wide applications. In the context of autonomous driving, encouraging results~\cite{PointRCNN, Yang_STD, pvrcnn} have been obtained with point cloud data from Light Detection and Ranging (LiDAR) devices.
Nonetheless, these single-modality LiDAR-based methods also have limitations: a typical LiDAR system can only perceive objects in a limited range~\cite{nuScenes}, and cannot distinguish semantic categories of similar structures, \eg, pedestrians and trees.

Meanwhile, imagery features, by nature, play a complementary role. It is believed that imagery features can facilitate more accurate detection by providing richer semantic information.
Much effort has geared towards adding RGB camera images to complement point cloud data~\cite{ChoSKR14, MMF_CVPR_2019, ContFuse_ECCV_2018, MVXNet, pointpainting}.
%
% For instance, Kalman filter has been used to perform a late fusion of point cloud and imagery measurements to enhance the robustness of detection. 
% %
% For example, Multimodal VoxelNet (MVX-Net)~\cite{MVXNet} performs point-wise fusion between features extracted from a 2D detector and LiDAR point cloud.
% The fused features are then voxelized into regular grid feature maps for 3D detection. 
%
However, top methods on the popular benchmarks\cite{nuScenes, KITTI} are mainly using single-modality.
Multi-modality approaches, despite having more information, seem to attain only marginal improvement.
The reason of this counter-intuitive phenomenon remains obscure.

% This paper seeks reasons behind the unexpectedly marginal improvement of multi-modality 3D object detection methods and shows that multi-modality augmentation plays pivotal roles.
% It remains obscure why multi-modality,  only supports a marginal improvement in 3D object detection.
%
This paper sets out to investigate why this is the case and finds it is the relatively insufficient data augmentation that make the performance of multi-modality algorithms under expectation.
Compared to recent point cloud methods, multi-modality methods typically use fewer kinds of augmentations, which is mainly due to the \emph{difficulties in maintaining the consistency} between point cloud and images.
In other words, multi-modality augmentations should ensure exact correspondence between a point in one modality and that of another modality after a series of transformations.
%
%that a point in one modality can find its exact corresponding point in another modality after the transformations.
% after the data of both modalities are augmented.
% However, some augmentations can not . Previous ignore them.
With such a constraint, previous methods~\cite{ContFuse_ECCV_2018, MVXNet} find difficulties in applying random flipping or rotations.
An alternative way for finding the correspondence is to use the inputs before augmentation.
However, it is not applicable to those points generated during model inference, \eg, box coordinates or votes~\cite{qi2019deep, ImVoteNet}.

% However, the first solution cannot allow some augmentations, \eg, random flipping, as it cannot be updated to the transformation matrix, and the second solution cannot find correspondence of points that are not from the input data (e.g., voxel centers or box centers).
% when conducting multi-modality data augmentation, a point from one modality must find its exact corresponding features in another modality and do the same transformation accordingly.
% \jiangmiao{State **why they cannot allow**, below statements are not clear enough. Just use **for example** to state what you said yesterday.}
% However, previous practices either cannot allow some augmentations to be applied (e.g., random flipping~\cite{ContFuse_ECCV_2018} or random rotation~\cite{MVXNet}),
% or 
%
% Multi-modality detectors use much fewer kinds of augmentation techniques than single-modality methods do,
% because multi-modality augmentations need to maintain the \emph{consistency} between multiple modalities while single-modality augmentations do not.
% The \emph{consistency} means that a point from one modality can find its correct corresponding features in another modality.
% Maintaining the \emph{consistency} is necessary to guarantee that multi-modality features can be correctly fused to facilitate more accurate detection.

To tackle this problem, we contribute a pipeline, named \emph{multi-modality transformation flow}, to ensure multi-modality consistency to cope with rich set of augmentations.
Multi-modality consistency can be maintained if and only if the augmentation can be reversed and replayed.
% \jiangmiao{Use one sentence to describe your priciple.}
% While previous methods~\cite{MVXNet, ContFuse_ECCV_2018} already used multi-modality augmentations,
% how to ensure consistency during augmentations has not been systematically discussed yet.
% the current practices lack a general method to allow various multi-modality augmentations to be applied while maintaining the \emph{consistency}.
Thus, \emph{transformation flow} records the parameters and orders of transformations used by each augmentation.
Then in the multi-modality fusion process, any point in the LiDAR coordinates could find its corresponding image pixel coordinates by reversing the point cloud transformations and replaying the image transformation.
In this pipeline, any transformation can be applied as long as it is invertible;
thus, one can now apply augmentations commonly found in single-modality detectors with equal effectiveness for multi-modality detectors (Fig.~\ref{fig:aug_compare}).

Based on \emph{multi-modality transformation flow},
we further propose a new augmentation approach, \emph{multi-modality cut and paste (MoCa)}.
State-of-the-art LiDAR-based detectors~\cite{PointRCNN, Yang_STD, pvrcnn} benefit significantly from cut and paste augmentation~\cite{SECOND}.
However, such an effective augmentation scheme is absent from multi-modality methods~\cite{MVXNet, pointpainting}, which severely limits their performance.
Single-modality cut and paste methods do not need to consider occlusions and physical plausibility in other modalities.
Consequently, they may construct scenes containing objects completely occluded in 2D images~\cite{SECOND} or paste an object at implausible locations.
These cases hinder the learning of multi-modality fusion modules as it makes the point from one modality fusing features from different objects in another modality.
% ; such omission s.
% \jiangmiao{State more specific about the challenge, current version is high level and do not have different with previous general statements.}
% \jiangmiao{The behaviors can be more clear.}
%
MoCa constrains paste operations to avoid occlusion between objects in both the bird's eye view (BEV) and the 2D imagery domain.
To further improve the detector's generalizability, we use randomly selected intersection over foreground (IoF) to determine the occlusion in imagery domain.
MoCa improves object detection performance by a large margin and it is readily applicable to many existing multi-modality detectors~\cite{MVXNet, pointpainting, ContFuse_ECCV_2018, fpointnet}.

% We further explore good practices in architecture design and optimization.
% The overall improvement brought by these practices and MoCa improves the MVX-Net~\cite{MVXNet}, a strong and generic multi-modality detector, by \textbf{11.3\%} in mAP on KITTI dataset (Fig.\ref{fig:aug_compare}).
%We further explore a pyramid aligned fusion module and a hybrid optimization strategy to improve the performance.
The transformation flow, MoCa, together with some beneficial practices, improve the mAP of MVX-Net~\cite{MVXNet}, a strong and generic multi-modality detector, by \textbf{11.3\%} moderate mAP on KITTI dataset (Fig.\ref{fig:aug_compare}) and \textbf{5.8\%} mAP on nuScenes dataset.
Without using an ensemble of class-specialized detectors, the enhanced MVX-Net achieves new state-of-the-art results on nuScenes dataset~\cite{nuScenes} and obtains competitive results on KITTI 3D benchmark~\cite{Geiger2012CVPR}.

% !TEX root = ../main.tex
\section{Related Work}

\noindent\textbf{3D object detection.}
Many approaches~\cite{Yang_STD, PointRCNN, SECOND, liu2019tanet} have been focusing on processing LiDAR point cloud to improve the performance of 3D object detection. 
% deep learning approaches for 2D object detection~\cite{lin2017_fpn, cascade_rcnn, Chen_2019_CVPR} improve drastically since Faster R-CNN~\cite{ren2015faster},
%
% while 3D object detection does not fall behind due to its wide applications.
% 
% Current state-of-the-art methods~\cite{PointPillars, PointRCNN, SECOND} are mainly LiDAR-based methods.
%
To deal with the irregular and unstructured nature of point cloud, common approaches either apply convolutional neural network (CNN) to the voxelized representation~\cite{VoxelNet, SECOND, PointPillars, liu2019tanet}
or process raw points~\cite{PointRCNN, qi2019deep, 3DSSD} by PointNets~\cite{Charles_2017, qi2017pointnet_plus}.
Recent methods~\cite{pvrcnn, Yang_STD, fast_point_rcnn} exploit both voxel representation and raw points.
There are also attempts~\cite{pseudo_lidar, 3DOP} that purely rely on cameras for 3D detection.

Previous works aggregate image and point cloud features from different views~\cite{MV3D, AVOD}; the efficiency is limited by the view aggregation for a large quantity of anchors~\cite{AVOD} or the proposals~\cite{MV3D}.
More recent works~\cite{MMF_CVPR_2019, ContFuse_ECCV_2018, MVXNet} fuse image features into each point, but they exhibit various feature misalignment issues.
For example, MVX-Net~\cite{MVXNet} quantizes image coordinates, while methods~\cite{MMF_CVPR_2019, ContFuse_ECCV_2018} based on ContFuse~\cite{ContFuse_ECCV_2018} use the nearest points for each BEV feature grid.
Frustum-based methods~\cite{fpointnet, fconvnet} obtain frustum proposals from an image and then apply PointNet~\cite{Charles_2017} to point cloud for 3D object localization.
Their performance is limited by the proposal qualities and they may not fully exploit the complementary information of multi-modalities.
ImVoteNet~\cite{ImVoteNet} skips the above-mentioned issue by fusing 2D votes in images and 3D votes in point clouds.
%
%We solve the misalignment issue in MVX-Net~\cite{MVXNet} by \textit{aligned feature fusion}, and enhance it by pyramid fusion with FPN~\cite{lin2017_fpn} to study the training strategies and data augmentation.

\noindent\textbf{Augmentations for detection.}
Data augmentation is crucial to improve the models' performance.
However, the current single-modality 3D object detection methods~\cite{Yang_STD, PointRCNN, pvrcnn} use more aggressive data augmentations than existing multi-modality methods~\cite{fpointnet, MMF_CVPR_2019, MVXNet} do.
Conventional image augmentations include but are not limited to random cropping, random flipping,
and multi-scale training~\cite{He_2016, mmdetection}.
For point cloud data, common augmentation techniques are random flipping, rotation, translation, and scaling~\cite{SECOND, VoxelNet, PointRCNN}.
In multi-modality 3D detection, the practices of augmentations vary across different methods~\cite{ContFuse_ECCV_2018, MMF_CVPR_2019, MVXNet}.
For example, ContFuse~\cite{ContFuse_ECCV_2018} skips random flip, 
whereas MVX-Net~\cite{MVXNet} does not apply random rotation and translation to maintain consistency between image and point cloud.
%
%The \emph{transformation flow} enables them, and their effectiveness are validated for multi-modality 3D object detection in this work.

%
There are augmentations applied to regions that contain objects of an image~\cite{cutout, cutmix, instaboost}.
A representative method is to cut and paste objects~\cite{Dwibedi_2017_ICCV, Dvornik_2018_ECCV, instaboost}.
While Dwibedi~\etal.~\cite{Dwibedi_2017_ICCV} paste objects randomly, recent works use a location probability map~\cite{instaboost}, semantic and depth information~\cite{GeorgakisMBK17}, or a visual context model~\cite{Dvornik_2018_ECCV} to guide the pasting process.
Cutting and pasting the points of objects is also common for LiDAR-based 3D detection methods~\cite{SECOND, liu2019tanet, Yang_STD} but is absent from multi-modality methods~\cite{fpointnet, MMF_CVPR_2019, MVXNet}.
%
%Therefore, we make the first attempt to enable \emph{multi-modality cut and paste (MoCa)} to bridge the gap in the augmentation techniques.
% !TEX root = ../main.tex
\section{Methodology}\label{sec:methods}

We are curious with the limited performance gain after extending single-modality to multi-modality 3D object detectors.
%
% We believe that a more careful investigation in the implementation of existing pipelines could shed lights for improving the performance of multi-modality detectors.
We find that multi-modality detectors use relatively fewer kinds of data augmentations than single-modality detectors do, due to the missing principle of maintaining multi-modality consistency in augmentation.
To this end, we first contribute a pipeline, named \emph{transformation flow}, to allow any invertible augmentations to be applied in multi-modality detection (Sec.~\ref{subsec:mm_aug}).
Then we propose \emph{multi-modality cut and paste (MoCa)} to further improve the performance (Sec.~\ref{subsec:moca}).
We choose MVX-Net~\cite{MVXNet} as a strong and generic baseline for our study.
Finally, we explore different architecture design and training strategies for better performance (Sec.~\ref{subsec:implement_detail}).
The findings can be easily extended to other multi-modality 3D object detectors.

%---------------------------------------------------------------
\subsection{Transformation Flow}\label{subsec:mm_aug}
%---------------------------------------------------------------
Data augmentation plays a pivotal role in improving the models generalizability.
%
% Overfitting is usually caused by insufficient training data due to difficulty in data collection and annotation.
% %
% For instance, there are fewer than 15K frames in KITTI 3D object detection dataset~\cite{Geiger2012CVPR}.
% This is far fewer than the conventional image datasets like COCO~\cite{lin2014coco} and ImageNet~\cite{ILSVRC15}, which contain about 328K and 1.46 billion annotated images, respectively.
%
% \noindent\textbf{Existing augmentation techniques.}
% We first carefully validate the gain of existing augmentation techniques on multi-modality data. Such experiments are new in the literature.
However, the augmentation strategies adopted by single-modality 3D detectors are more aggressive than those used by multi-modality methods.
For example, global rotation and random flip are widely applied by single-modality methods~\cite{PointRCNN, Yang_STD, liu2019tanet} but are absent from some multi-modality methods~\cite{MVXNet,ContFuse_ECCV_2018, MMF_CVPR_2019} due to the \emph{difficulties in maintaining the consistency} between point cloud and image data.
% For consistency, the correct correspondence between points and image pixels is maintained and that the points can be successfully fused with their corresponding image features. 

\noindent\textbf{Multi-modality consistency.}
The essence of maintaining multi-modality \emph{consistency} during augmentation is to maintain the correct correspondence between points and image pixels;
thus, the points can still be correctly fused with their corresponding image features after a series of augmentations. %ensuring point cloud and its corresponding image features are fused correctly.
While previous studies~\cite{MVXNet, ContFuse_ECCV_2018} have applied augmentation in multi-modality detection,
they use much fewer augmentations than those single-modality methods and the issue of maintaining consistency across different modalities has not been systematically studied.
Here, we contribute a pipeline named \emph{multi-modality transformation flow}, which is useful for maintaining the multi-modality consistency during augmentation, enabling more aggressive augmentation strategies for multi-modality detection.

\begin{figure}[t]
  \centering\includegraphics[width=0.47\textwidth]{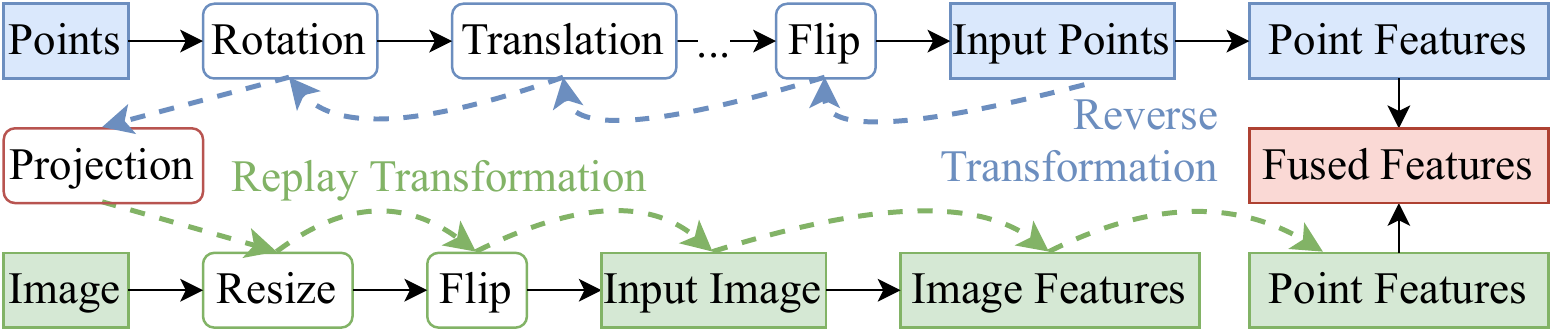}
  \vspace{-6pt}
  \caption{\small{\textbf{Multi-modality transformation flow.} A point in point cloud data finds its image coordinate by reversing point cloud augmentation and replaying image augmentation.
  }}\label{fig:points_transformation}
\vspace{-15pt}
\end{figure}

\noindent\textbf{Multi-modality transformation flow.}
As shown in Fig.~\ref{fig:points_transformation}, the multi-modality transformation flow records all the transformations of point cloud and image data during data augmentations.
Such transformation flow is required to transform the augmented data back for finding the correct correspondence between the point cloud and image pixels during fusion.
Most augmentations are reversible, \ie, they contain a forward transformation to augment the data, with a reverse transformation to transform the data back into its original state.
Before training, the image and point cloud data are augmented by different augmentations independently.
Note that the transformations of points are equivalent to transforming the LiDAR sensor to a new position, resulting in new point coordinates but not affecting the captured image as the camera is not transformed. 
Consequently, rotation and translation can be applied to point cloud data, but they do not need to be applied to the image simultaneously.

\noindent\textbf{Reverse and Replay.}
With transformation flow, a point in one modality can obtain its corresponding point in another modality by reversing the augmentations of its own modality and replaying the augmentation of the other modality (Fig.~\ref{fig:points_transformation}).
Specifically, during fusion, the reverse transformation of each augmentation is applied to the augmented points following the inverse order of point cloud augmentations (reverse).
Next, we can safely project the points onto the imagery pixel coordinates using the calibration information of data~\cite{KITTI, nuScenes}.
The projected points then obtain their corresponding image features after going through the forward transformations following the same order of the corresponding image augmentations (replay).

\noindent\textbf{Applications.}
With \emph{multi-modality transformation flow}, any augmentation, as long as it is reversible, can be used to augment multi-modality data without sacrificing the consistency.
Therefore, now we can revisit and validate the gain of existing augmentation techniques that can be extended from single-modality to multi-modality augmentations.
We compare these augmentation techniques step by step (Fig.~\ref{fig:aug_compare}); such experiments are new in the literature.
% We carefully  existing augmentation techniques on multi-modality data. 
The results show that global flipping, scaling, rotation, and translation are all essential augmentations to improve a multi-modality detector, enabled by the transformation flow.

Some methods~\cite{SECOND, VoxelNet} apply a small amount of noise (\eg, random translation and rotation) to each ground truth object separately.
They can also be applied in \emph{multi-modality transformation flow} by recording the transformation of each single ground truth objects.
However, recent works suggest that such strategy either does not add value~\cite{pvrcnn} or hurts the performance in some scenarios~\cite{PointPillars}.
% Therefore, we leave it to future research.

%---------------------------------------------------------------
\subsection{Multi-modality Cut and Paste}\label{subsec:moca}
%---------------------------------------------------------------
\begin{figure*}[t]
  \centering\includegraphics[width=\textwidth]{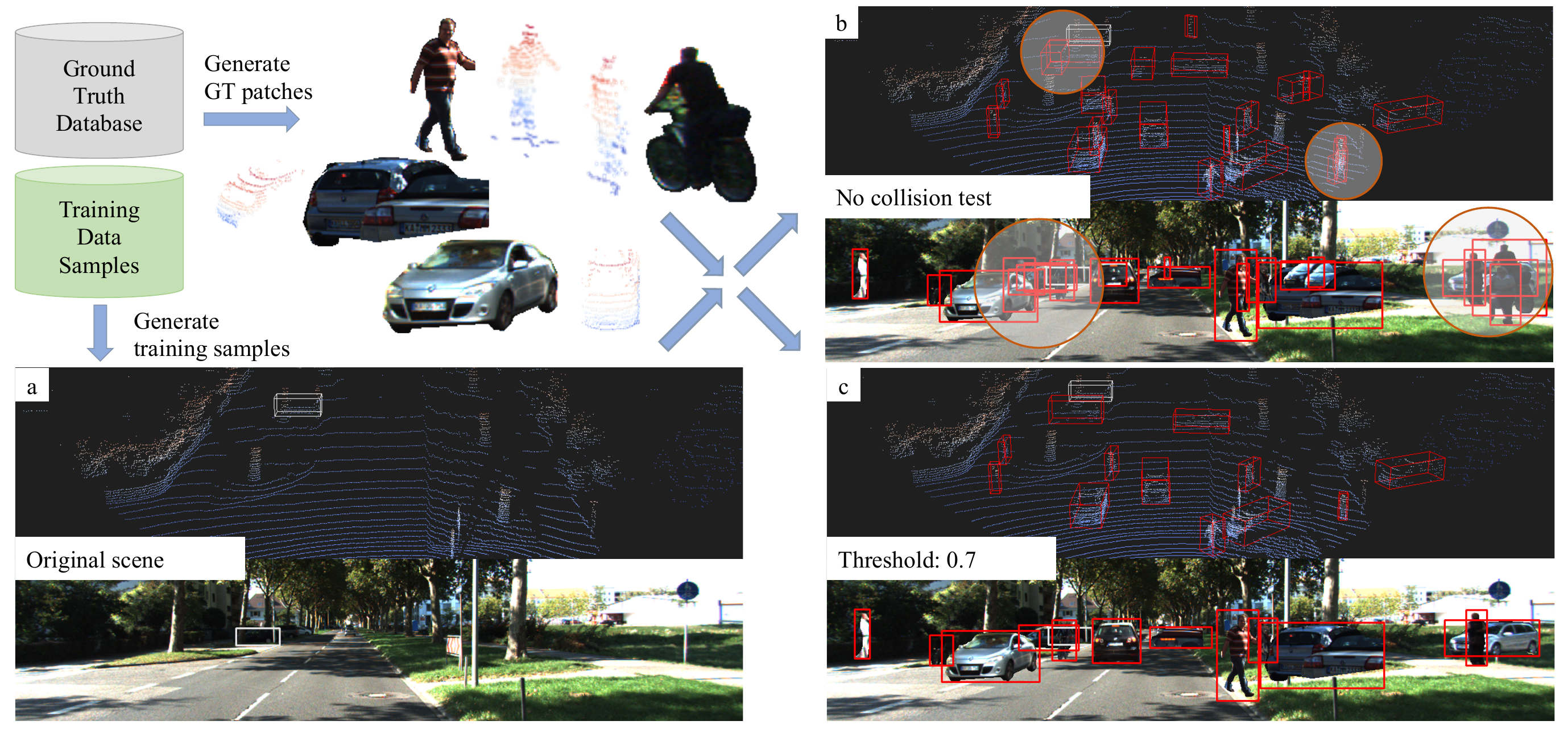}
  \vspace{-21pt}
  \caption{\small{\textbf{Multi-modality cut and paste.}
  Given one training frame and some GT objects generated by the database, 
  MoCa checks collisions in both BEV and 2D image based on Intersection over Foreground (IoF).
  It then pastes the valid patches of objects to all the modalities.
  A blind pasting operation will cause many heavily overlapped objects or put objects at implausible positions (circled area in (b)), which is far from the natural distribution of data.
  The gray and red bounding boxes indicate the original object in the current frame and the pasted objects, respectively.
  The original frame (a) contains very few objects and is enriched with more objects after multi-modality cut and paste (c). %\cavan{We label 1, 2, ... 4 in the figure but they are never referred to. The blue box can hardly be seen.}
  The figure is best seen in color.
  }}\label{fig:mmaug_sample}
	\vspace{-15pt}
\end{figure*}

% During the comparison between augmentations for multi-modality and single-modality detectors, we find that cut and paste is one of the major missing augmentation techniques, as shown in Fig.~\ref{fig:aug_compare}.
%Cut and paste, which is widely applied for different tasks including object detection~\cite{Dvornik_2018_ECCV}, instance segmentation~\cite{instaboost}, and instance detection~\cite{Dwibedi_2017_ICCV},
Cut and paste is an effective strategy to create a diverse combination of scenes and objects when data is limited~\cite{Dvornik_2018_ECCV,instaboost,Dwibedi_2017_ICCV}. 
It is a common augmentation technique in single-modality detectors but not in multi-modality detectors.
% Cut and paste is a straightforward strategy to create diverse combination of scenes and objects when data is limited,
% which is widely applied for different tasks including object detection~\cite{Dvornik_2018_ECCV}, instance segmentation~\cite{instaboost}, and instance detection~\cite{Dwibedi_2017_ICCV}.
% 
Recently, SECOND~\cite{SECOND} introduces cut and paste into the point cloud domain, named as ground truth sampling (GT-sampling).
It is shown that GT-sampling not only accelerates model convergence but also reduces class imbalance issues.
Therefore, almost all state-of-the-art single-modality 3D object detectors~\cite{PointRCNN, liu2019tanet, Yang_STD, pvrcnn, 3DSSD} adopt GT-sampling to boost their performance.

It is non-trivial to consistently extend GT-sampling to the imagery domain in multi-modality methods.
%
%This is because point cloud data is relatively sparse and irregular compared to imagery pixels.
A point cloud patch that is visible from the bird's eye view (BEV) does not guarantee its corresponding image patch is also perceivable in the imagery domain. Often, an object may be occluded in the image plane and thus its imagery content only captures the features of the occluding object.
A blind cut and paste would risk having inconsistent point cloud and imagery patches.
To circumvent this intricate problem, multi-modality methods~\cite{MMF_CVPR_2019, MVXNet} resort to a lower starting baseline without using GT-sampling. %, but they still need to compare performance with single-modality methods~\cite{PointRCNN, Yang_STD} that use GT-sampling.

To our knowledge, this is the first work that investigates underlying challenges in multi-modality cut and paste for 3D object detection. We further propose a more general form of GT-sampling named \emph{multi-modality cut and paste (MoCa)}.
%
% It brings large improvement and it is widely applicable to existing multi-modality methods.
It is readily applicable to existing multi-modality methods~\cite{MVXNet, pointpainting, ContFuse_ECCV_2018, fpointnet} and brings substantial improvement.

% \vspace{-2pt}
MoCa first builds a ground truth database for each annotated object offline. 
Specifically, point cloud for each object and its corresponding image patch is cropped before training, using ground truth 3D bounding boxes and 2D masks, respectively.
%
% During training, our method will sample some ground truth objects from the database and paste their corresponding point cloud and image patches to the current frame.
%
During training, MoCa randomly samples point cloud-image patch pairs and paste them to the original scene according to their 3D bounding boxes and 2D masks (Fig.~\ref{fig:mmaug_sample}).
To avoid boundary artifacts caused by image patches, we follow \cite{Dwibedi_2017_ICCV} to apply random blending to smoothen the boundaries of image patches.
Such an operation is not needed for point cloud since the data is sparse.

% \vspace{-1pt}
\noindent\textbf{Occlusion handling.}
The non-trivial part of multi-modality cut and paste lies in occlusion handling.
Image-based cut and paste~\cite{Dwibedi_2017_ICCV, Dvornik_2018_ECCV} usually pastes objects at different locations in the image and ignores the physical plausibility.
On the other hand, point cloud cut and paste~\cite{SECOND} only avoids occlusion in BEV because objects are generally assumed to be on the same ground plane and well separated in BEV. % \cavan{not sure what this sentence means, rephrase.}
Potential occlusions in 2D image are neglected because current 3D object detectors~\cite{SECOND, Yang_STD, PointPillars} usually predict bounding boxes only from BEV.
However, during the multi-modality fusion process, due to the occlusion, the projected points of an occluded object might obtain the image features of occluding objects (Fig.~\ref{fig:mmaug_sample} (b)).
This makes the image features ambiguous and increases the difficulty in training feature extractors.
Therefore, not handling the occlusion in 2D image will affect the overall performance as validated by our experiments (Table~\ref{tab:mmaug_thr}). %\cavan{I use the word quality here, not sure if this is correct. I still don't get why occlusion in 2D matters}
% \wenwei{Our experiment (Table~\ref{tab:mmaug_thr}) shows that not handling the 2D occlusion will degrade the performance by 1.4, 1.1, and 2.3 mAP on easy, moderate, and hard conditions, respectively, over all the categories.}
% \cavan{not sure what this sentence means too. Why it doesn't handle potential occlusion in 2D image? Also we need to say what happen if occlusion is not handled so that the readers can appreciate the severity. How much drop in performance?}.

MoCa considers the consistency in both point cloud and image modalities.
Specifically, given a batch of objects with their point cloud and their corresponding image patches,
the multi-modality cut and paste first discards overlapped objects in BEV and then carefully handles the occlusion in 2D images. %\zwang{what does occluded objects mean here? In my understanding, occluded means sth is in front of the occluded objects so that the latter cannot be observed. Here do you mean the two objects may overlap with each other in BEV? If so, maybe use another word instead of occluded}
Given a set $P = \left\{p_i | i = 1, 2, ..., N\right\}$ containing the original objects and the objects to be pasted, %\zwang{I am not sure whether this object set includes both original objects in this frame and the gt database sampled objects. Need to clarify.} 
we use Intersection-over-Foreground (IoF) to represent the occlusion degree of an object $p_i$ in the 2D image as
\begin{equation}
  IoF(p_i, P) = \max\left\{\frac{p_i \bigcap p_j}{p_i} |j \neq i\right\}. \nonumber
\end{equation}
Once a sampled object's IoF is greater than a given threshold or that object makes any one of the original boxes' IoF greater than the given threshold, the sampled object will not be pasted in the current training iteration.
The original objects will not be discarded.

\noindent\textbf{Mixed IoF thresholds.}
Different IoF thresholds lead to a different number of objects being pasted.
Therefore, we propose to use a mix of occlusion thresholds to provide more diverse occlusion cases and scenes during training to improve the detector's robustness and generalization ability.
Specifically, given a threshold set (we use $\left\{0, 0.3, 0.5, 0.7\right\}$ in this work), a threshold will be randomly chosen from the set.
Then objects whose occlusion degrees (in the batch) are greater than the threshold will be discarded during each iteration.
The remaining objects' point cloud and image patches will then be pasted to the positions specified by their 3D and 2D bounding boxes \emph{without random perturbation} to ensure consistency, respectively. 
Image patches are pasted in the order of their depth, \ie, the farther the object, the earlier it is pasted.

\begin{figure}[t]
    \centering
    \includegraphics[width=0.45\textwidth]{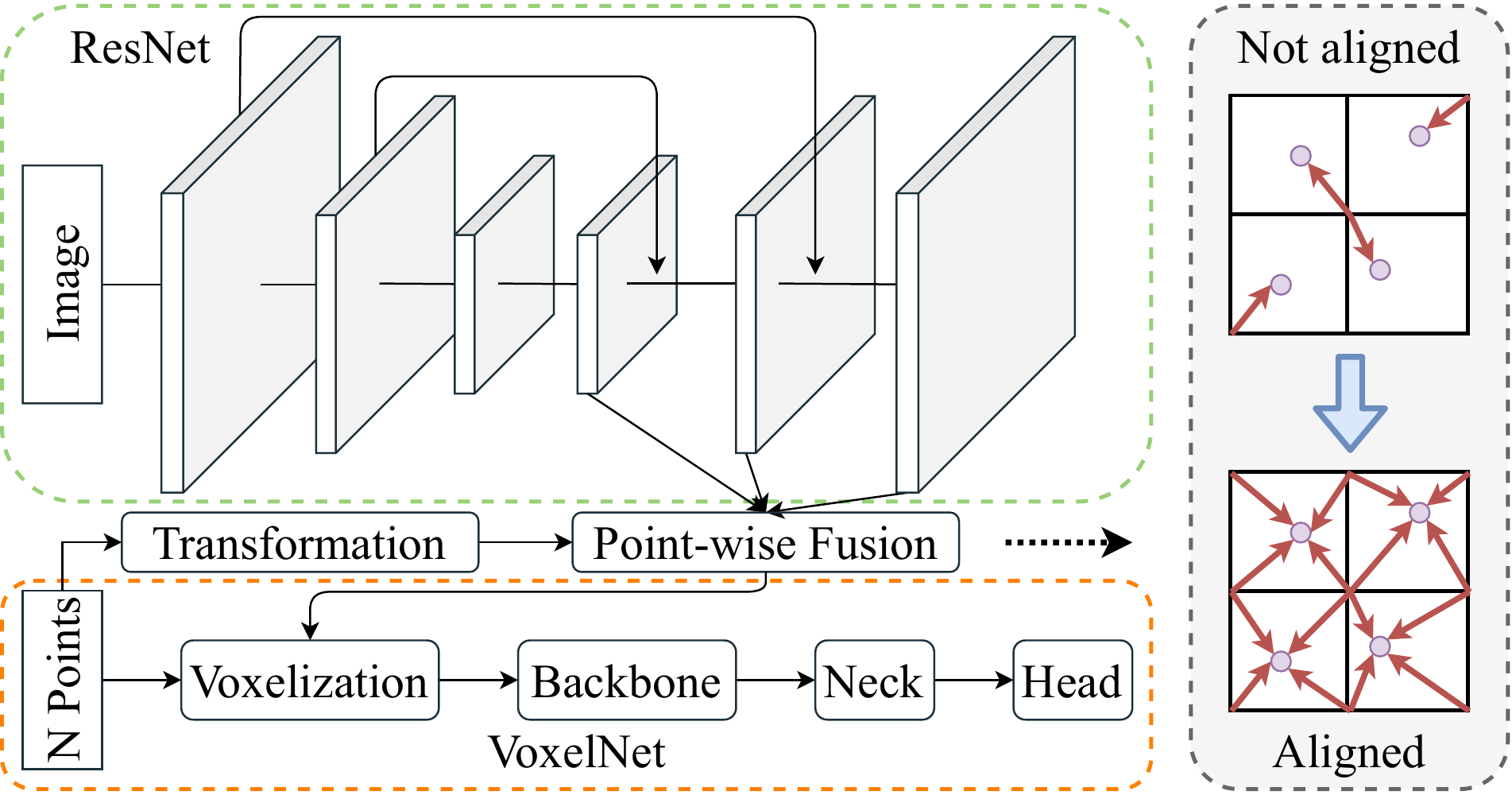}
    \vspace{-6pt}
    \caption{\small{Enhanced MVX-Net. We fuse FPN features with point cloud features using not-quantized coordinate.
    The fused features are voxelized into regular grid feature maps, which are then fed to VoxelNets (\eg, PointPillars~\cite{PointPillars}) for 3D detection.}}\label{fig:mvx_design}
    \vspace{-18pt}
\end{figure}

%---------------------------------------------------------------
\subsection{Exploring Beneficial Practices}\label{subsec:implement_detail}
%---------------------------------------------------------------
% \noindent\textbf{Enhanced MVX-Net.}
In this paper, we use MVX-Net~\cite{MVXNet} to study multi-modality augmentation as it is simple and generic.
MVX-Net uses a pre-trained Faster R-CNN as an image feature extractor and adopts VoxelNets (\eg, PointPillars~\cite{PointPillars} or SECOND~\cite{SECOND}) for 3D detection.
During implementation, we explore some beneficial practices in architecture design and optimization.

\noindent\textbf{Aligned pyramid feature fusion.}
For feature fusion, the point cloud is first transformed from LiDAR coordinates to camera coordinates and then projected to the image pixel coordinates\footnote{Details are provided in the appendix.}.
Once the pixel coordinates $P_{img}$ are obtained, MVX-Net selects image features using the quantized coordinates $P_{img}'$ and concatenate them with the point cloud feature.
However, as shown in Fig.~\ref{fig:mvx_design}, the quantization introduces feature misalignment in the fusion process since the quantized coordinates are not an accurate projection from the given points.
Such misalignment brings adverse effects in multi-modality detection because the quantization is applied to at least 10K projected points with their corresponding features.
Inspired by RoIAlign~\cite{mask_rcnn}, we introduce \textit{aligned feature fusion}, which uses differentiable bilinear sampling kernel~\cite{STN} to overcome the misalignment issue and obtain the feature of a given point as follows:
\begin{equation}
    \sum_{H}^{n}\sum_{W}^{m}U_{nm}\max(0, 1-|P_{x} - m|)\max(0, 1-|P_{y} - n|), \nonumber
\end{equation}
% \cavan{what is $H$, $W$, $U$, $P_x$ and $P_y$?}
where $H$ and $W$ are the height and width, respectively, of the feature map $U$, and $P_{img}= (P_{x}, P_{y})$ is the pixel coordinates of a projected point.

We further enhance the MVX-Net~\cite{MVXNet} by adding a FPN~\cite{lin2017_fpn} in image branch (Fig.~\ref{fig:mvx_design}).
For the feature map in each scale, the pixel coordinate $P_{img}$ is divided by the strides of the current feature map and then used to select the corresponding image features.
Then we obtain the multi-scale imagery features of each point from different scales of feature maps, concatenate them together, and fuse them with the points' point cloud features by a linear layer.

\begin{figure}[t]
    \centering
    \includegraphics[width=0.4\textwidth]{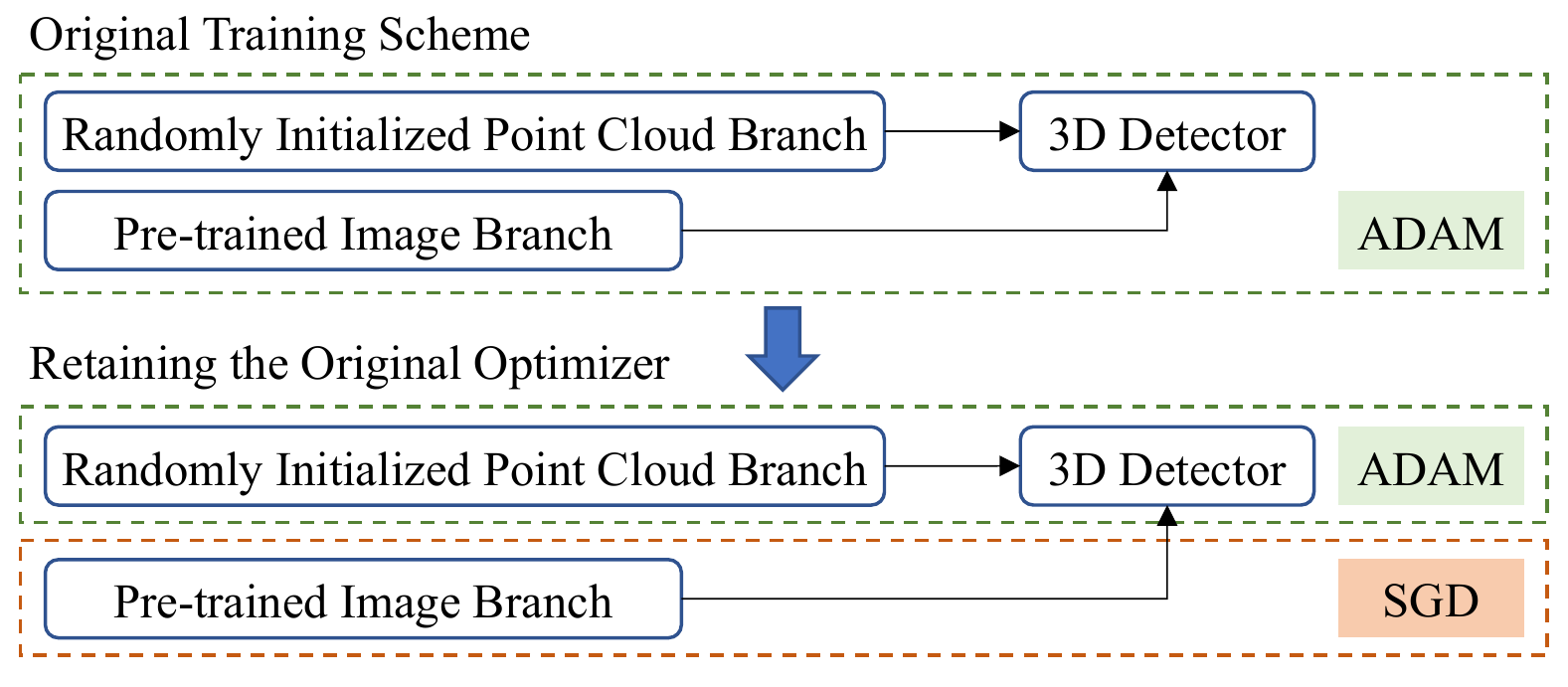}
    \vspace{-9pt}
    \caption{\small{Retaining the original optimizer for each modality is beneficial.}}\label{fig:cocktail_optim}
	\vspace{-12pt}
\end{figure}

\noindent\textbf{Retaining the original optimizers.}
Multi-modality 3D object detectors~\cite{ContFuse_ECCV_2018, fpointnet, MVXNet, MMF_CVPR_2019} usually use a pre-trained model for image feature extraction because there are limited image data in the problem domain.
The image feature extractor is typically pre-trained on 2D recognition tasks~\cite{ILSVRC15, lin2014coco} using an SGD optimizer to ensure good performance. On the other hand, ADAM is shown superior in handling irregular and unstructured point cloud data~\cite{Charles_2017, qi2017pointnet_plus}.
%
% A common practice is then to train the point cloud feature extractor from scratch jointly with the pre-trained image feature extractor for 3D object detection with an ADAM optimizer.
Hence, for multi-modality 3D object detection, a common practice is to train the point cloud feature extractor from scratch jointly with the pre-trained image feature extractor using an ADAM optimizer.
Empirically, we observe that the switch of optimizer from SGD to ADAM in the image branch causes a slight performance drop.
%
% Measured by Centered Kernel Alignment (CKA)~\cite{CKA}, the knowledge deviation after the switch can go up to 99\%.
We ameliorate this problem by retaining the original optimizer of each modality (Fig.~\ref{fig:cocktail_optim}).
Specifically, we retain the use of a SGD optimizer for the image feature extractor and an ADAM optimizer for the point cloud branch  during the joint training. We ablate the effect in the experiments.
%
% Due to the differences in modality and representation ability between the image branch and the point cloud branch, retaining the original optimizers may update the image branch at a slower speed, 
% \eg, updates the parameters once after four training iterations.
%
%Such a choice of using different optimizer for each modality may be explained by common practices in each individual modality -- existing methods usually use SGD for image recognition tasks~\cite{ren2015faster,Chen_2019_CVPR,He_2016} while ADAM is more commonly used for point cloud representation learning because ADAM is shown superior in handling irregular and unstructured point cloud data~\cite{Charles_2017, qi2017pointnet_plus}.

% !TEX root = ../main.tex
\begin{table*}[ht]
	\caption{
        \small{Comparison with published multi-modality methods on KITTI 3D test benchmark.
        `Ensembled' indicates whether the results are ensembled by class-specialized detectors. The best results are bolded
        }
	}\label{tab:mvx_test_3d}
	\vspace{-21pt}
    \begin{center}
        \addtolength\tabcolsep{-0.3em}
		\scalebox{0.86}{\begin{tabular}{c|c|c|c|c|c|c|c|c|c|c|c|c|c}
        \multirow{2}{*}{Method}&\multirow{2}{*}{Ensembled} & \multicolumn{3}{c|}{mAP (\%)} & \multicolumn{3}{c|}{Pedestrian}  & \multicolumn{3}{c|}{Cyclist} & \multicolumn{3}{c}{Car}\\ \cline{3-14}
            && Easy & Mod. & Hard & Easy & Mod. & Hard & Easy & Mod. & Hard & Easy & Mod. & Hard  \\ \hline
            % PointPillars~\cite{PointPillars}&\multirow{4}{*}{L}&70.4&	58.3&	53.3	&51.5	&41.9	&38.9	&\newgreen{77.1}	&58.7	&51.9	&82.6	&74.3	&69.0 \\
            % PointRCNN~\cite{PointRCNN}	&&70.0&	57.9&	53.1	&48.0	&39.4	&36.0	&75.0	&58.8	&52.5	&\newgreen{87.0}	&75.6	&\cyan{70.7} \\
            % TANet~\cite{liu2019tanet}	&&\newgreen{71.3}	&59.9	&53.9	& \red{53.7}	&\red{44.3}	&\red{40.5}	&75.7	&59.4	&52.5	&84.4	&\newgreen{75.9}	&68.8 \\
            % STD~\cite{Yang_STD}&	        &\cyan{73.3}&  \cyan{61.3}&	\red{56.2}	&\cyan{53.3}	&42.5	&38.4	&\cyan{78.7}	&\cyan{61.6}	&\cyan{55.3}&\red{88.0}	&\red{79.7}	&\red{75.1} \\\hline
            AVOD-FPN~\cite{AVOD}&\multirow{4}{*}{$\surd$}&65.8&	54.9&	49.9	&50.5	&42.3	&39.0&63.8	&50.6	&44.9	&83.1	&71.8	&65.7 \\
            F-PointNet~\cite{fpointnet}&	&68.3&	56.0&	49.2	&50.5	&42.2	&38.1	&72.3	&56.1	&49.0	&82.2	&69.8	&60.6 \\
            PointPainting~\cite{pointpainting}&	&70.0   &58.8&	53.6&	50.3&	41.0&	37.9&	77.6&	63.8&	55.9&	82.1&	71.7&67.0 \\
            F-ConvNet~\cite{fconvnet}&	&\textbf{73.8}&\textbf{61.6}&54.0	&\textbf{52.2}	&43.4	&38.8	&\textbf{82.0}&\textbf{65.1}&\textbf{56.5}&\textbf{87.4}&\textbf{76.4}&66.7\\ \hline
            MoCa (ours)	&$\times$&71.0&	60.2&\textbf{54.7}&50.9	&\textbf{43.7}&\textbf{40.0}&76.1	&61.0	&53.4	&86.0	&75.9	&\textbf{70.7}\\
		\end{tabular}}
	\end{center}
	\vspace{-21pt}
\end{table*}

\begin{table*}[ht]
	\caption{
        \small{Comparison with previous methods on nuScenes validation set.
        `Con. Veh.', `Ped.', and `T.C.' are the abbreviations of construction vehicle, pedestrian, and traffic cone, respectively.
        `FA' means FreeAnchor~\cite{FreeAnchor} and `3$\times$' means longer training schedule.
        NDS score, mAP, and APs of each categories are reported. The single class AP not reported in the paper is marked by `-'.
        The best results are bolded
        }
	}\label{tab:mvx_nus_val}
	\vspace{-18pt}
    \begin{center}
        \addtolength\tabcolsep{-0.3em}
		\scalebox{0.85}{\begin{tabular}{c|c|c|c|c|c|c|c|c|c|c|c|c|c}
            Method &Modality& NDS (\%) &mAP (\%) &Car& Truck & Bus & Trailer & Con. Veh. &Ped. & Motor & Bicycle  & T.C. & Barrier \\ \hline
            PointPillars~\cite{PointPillars}&L&46.8&28.2&75.5&31.6&44.9&23.7&4.0&49.6&14.6&0.4&8.0&30.0 \\
            3D-CVF~\cite{3DCVF}&L + I&49.8&42.2&79.7&37.9&55.0&36.3&-&71.3&37.2&-&40.8 &47.1 \\
            PointPillars~\cite{PointPillars}+FPN~\cite{lin2017_fpn}&L&53.4&40.1&80.6&35.9&43.5&29.2&5.4&71.9&34.9&11.8&35&52.6 \\
            3DSSD~\cite{3DSSD}&L &56.4&42.6&81.2 & 47.2 & 61.4 & 30.5 &12.6 & 70.2 & 36.0 & 8.6 & 31.1 & 47.9 \\
            PointPainting~\cite{pointpainting}&L + I &58.1&46.4&77.9&35.8&36.1&37.3&15.8&73.3&41.5&24.1&62.4&60.2 \\
            CenterPoint~\cite{CenterPoint}&L&65.0&56.6&84.6&54.7&66&32.3&15.1&84.5&56.9&38.6&67.4&66.1 \\ \hline
            MoCa & L+I&58.1&47.9&82.4&41.5&49.6&28.6&9.1&79.1&50.3&27.2&49&61.9 \\ \hline
            \multicolumn{14}{l}{\quad\emph{MoCa with more improvements}} \\ \hline
            + FA~\cite{FreeAnchor}    & \multirow{5}{*}{L + I}&60.3&52.9&83.6&48.5&56.4&31.4&10.8&81.6&61.0&35.7&58.1&61.7 \\
            + FA + RegNetX-400MF\cite{regnet}  &&62.1&55.2&84.2&51.7&63.6&34.2&18.0&82.0&61.8&32.9&59.4&64.0 \\
            + FA + RegNetX-1.6GF &&64.7&	58.4& 85.2&55.5&64.5&35.0&20.8&84.5&68.0&42.8&62.3&65.1 \\
            + FA + RegNetX-3.2GF &&65.4&59.2&85.7&56.8&66.2&35.8&21.7&84.4&67.4&44.2&62.8&66.3 \\
            + FA + RegNetX-3.2GF + 3$\times$~ &&\textbf{67.8}&\textbf{62.5}&\textbf{87.0}&\textbf{60.8}&\textbf{68.2}&\textbf{39.7}&\textbf{23.9}&\textbf{85.8}&\textbf{70.2}&\textbf{51.6}&\textbf{68.1}&\textbf{69.5}	 \\
		\end{tabular}}
	\end{center}
	\vspace{-24pt}
\end{table*}

\section{Experiments}\label{sec:Experiments}
%---------------------------------------------------------------
%\subsection{Dataset Details}
%---------------------------------------------------------------
We validate our framework on the KITTI dataset~\cite{Geiger2012CVPR}, and nuScenes dataset~\cite{nuScenes}.

\noindent\textbf{KITTI dataset.}
The KITTI dataset contains 7481 training images and 7518 test images, both with their corresponding point cloud. 
We train all the models using the train split containing 3712 samples and evaluate them on the validation split consisting of 3769 samples, 
following previous works~\cite{MV3D, SECOND, PointRCNN}.
The KITTI benchmark evaluates the models by Average Precision (AP) of each class (car, pedestrian, and cyclist) under easy, moderate, and hard conditions.
For simplicity, we use mean AP over three classes to measure overall performance of the models in the ablation study.

\noindent\textbf{nuScenes dataset.}
The nuScenes dataset contains 28130 synchronized multi-view images and point cloud samples for training, 
6019 samples for validation, and 6008 samples for test benchmark.
The dataset contains 10 categories and we evaluate the models on these 10 classes by NDS metric~\cite{nuScenes}.
The NDS metric not only measures the mean AP but also takes translation, scale, orientation, velocity, and attribute errors of the true positives into considerations.

%---------------------------------------------------------------
\subsection{Benchmark Results}
%---------------------------------------------------------------
\noindent\textbf{KITTI dataset.}
Then we compare our method with other published methods on the KITTI 3D detection benchmark for completeness.
Note that many previous works~\cite{Yang_STD, SECOND, PointPillars, AVOD, fconvnet, liu2019tanet} train \textbf{specialized models with different hyperparameters for different categories and ensemble their results on the benchmark}.
However, using multiple detectors for multiple classes is not ideal for real-world applications.
Therefore, in this work, we train all the models including all baselines on all three classes without tuning the models for specific categories.
Table~\ref{tab:mvx_test_3d} shows that MoCa achieves promising performance among multi-modality methods. 
Despite a single model for all three classes, MoCa achieves competitive performance against other baselines that use an ensemble of class-specific detectors.
On hard conditions, only MoCa obtains top-3 results of all the categories.
In comparison, other methods do not exhibit such generalizability.

\noindent\textbf{nuScenes dataset.}
% Table 8 写出来 ensemble & not ensemble.
We compare our method with other published methods on the validation set of nuScenes dataset~\cite{nuScenes} in Table~\ref{tab:mvx_nus_val}.
The dataset has more diverse scenes and need to detect 10 categories, which is more challenging.
We report the performance of enhanced MVX-Net with MoCa, using the image branch from HTC pre-trained on nuImages dataset.
The results show that MoCa surpasses PointPainting~\cite{pointpainting}, the previous multi-modality state of the art, by \textbf{1.5\%} mAP.
To evaluate the scalability of our methods and compare with methods using large model~\cite{CBGS}, we also report the performance of MoCa enhanced by FreeAnchor~\cite{FreeAnchor}, RegNetX~\cite{regnet}, and longer schedule with stronger augmentations.
MoCa achieves new state-of-the-art results not only on the overall metric, but also on the AP of all the categories.
The final performance of MoCa surparsses the previous best result achieved by a large model of CenterPoint~\cite{CenterPoint} trained by CBGS~\cite{CBGS}, with an absolute improvement of \textbf{2.8\%} NDS and \textbf{5.9\%} mAP.
% The results show that MoCa surpasses previous methods~\cite{CenterPoint} by a large margin and achieve the new state-of-the-art performance step by step.

\begin{table*}[h]
	\caption{
        \small{Comparison of different augmentation strategies for single-modality 3D detector SECOND~\cite{SECOND} 
        and multi-modality 3D detector MVX-Net~\cite{MVXNet} on KITTI.
        The augmentation technique is in order, whereby, each augmentation is added onto the previous ones sequentially
        }
	}\label{tab:mmaug_add}
	\vspace{-30pt}
	\begin{center}
		\scalebox{0.85}{\begin{tabular}{c|c|c|c|c|c|c|c|c|c|c|c}
            \multirow{2}{*}{Method} & \multirow{2}{*}{Augmentation} & mAP (\%) & \multicolumn{3}{c|}{Pedestrian}  & \multicolumn{3}{c|}{Cyclist} & \multicolumn{3}{c}{Car}  \\ \cline{3-12}
            &  & Mod. & Easy & Mod. & Hard & Easy & Mod. & Hard & Easy & Mod. & Hard  \\ \hline
            \multirow{6}{*}{SECOND} & no augmentation & 47.4 & 46.5	&40.0	&34.7	&55.3	&38.7	&32.8	&71.5	&63.4	&63.4\\
            & + flipping &51.9 & 57.6 & 50.9 & 45.1	&53.7	&37.6	&37.2	&77.3	&67.3	&65.8 \\
            & + scaling &56.5 & 54.8 & 47.0 & 44.5	&67.2	&48.4	&42.3	&84.5	&74.1	&67.6 \\
            & + rotation &59.3 & 59.8 & 53.0 & 46.6	&66.5	&48.8	&48.0	&87.1	&76.0	&68.9 \\
            & + translation &60.7 & 61.0 & 54.6 & 51.9	&69.0	&50.7	&49.2	&87.2	&76.8	&75.6 \\
            & + cut and paste &68.1 & 68.1 & 61.1 & 54.0	&79.4	&65.8	&60.8	&87.5	&77.4	&75.5 \\ \hline
            \multirow{6}{*}{MVXNet} & no augmentation & 48.5 & 49.1	&45.5	&40.5	&53.2	&35.4	&31.2	&75.3	&64.5	&57.8\\
            & + flipping & 53.6 & 49.7	&43.3	&41.7	&57.2	&45.6	&40.4	&81.5	&71.9	&65.9\\
            & + scaling & 56.6 & 54.4	&47.0	&44.6	&66.2	&49.5	&42.9	&82.7	&73.4	&67.1\\
            & + rotation & 60.1 & 62.7	&54.7	&52.8	&67.2	&49.5	&42.6	&87.7	&76.0	&68.8\\
            & + translation & 60.5 & 62.7	&55.6	&53.2	&65.7	&49.1	&48.4	&87.2	&76.9	&75.5\\
            & \textbf{+ MoCa (ours)} & \textbf{70.2} & \textbf{68.6}	&\textbf{61.9}	&\textbf{54.7}	&\textbf{86.0}	&\textbf{71.2}	&\textbf{65.0}	&\textbf{87.9}	&\textbf{77.6}	&\textbf{76.0}\\
		\end{tabular}}
	\end{center}
	\vspace{-27pt}
\end{table*}

\begin{table}[t]
    \centering
    \caption{\small{Comparison of different IoF thresholds. 
    `No test' means not applying collision test, and `mixed' means using different thresholds during training}}\label{tab:mmaug_thr}
    \vspace{-9pt}
    \addtolength\tabcolsep{-0.3em}
    \subfloat[Multi-modality 3D detector\vspace{-6pt}]{\scalebox{0.85}{\begin{tabular}{c|c|c|c}
        \multirow{2}{*}{Threshold} & \multicolumn{3}{c}{3D mAP (\%)}\\ \cline{2-4}
        &Easy & Mod. & Hard \\ \hline
        No test &79.5&69.5&66.0  \\ \hline
        0       &81.8&69.4&64.9 \\
        0.3     &80.9&69.7&66.0 \\
        0.5     &80.0&69.6&\textbf{66.0} \\
        0.7     &80.0&69.7&65.6 \\ \hline
        Mixed   &\textbf{80.9} &\textbf{70.2}&65.2  \\
   \end{tabular}}\hspace{3mm}}\hspace{3mm}
   \subfloat[2D detector\vspace{-6pt}]{\scalebox{0.85}{\begin{tabular}{c|c|c|c}
    \multirow{2}{*}{Threshold} & \multicolumn{3}{c}{2D mAP (\%)}\\ \cline{2-4}
    &Easy & Mod. & Hard \\ \hline
    No test  & 84.9  & 78.2 & 72.1 \\ \hline
    0   & \textbf{87.0}& 78.5	& 73.3 \\
    0.3 & 86.1	& 79.0	& 73.9 \\
    0.5 & 85.8 & 79.0 & 73.6 \\
    0.7 & 86.8	& 78.7	& 73.2 \\ \hline
    Mixed  & 86.3 & \textbf{79.3} & \textbf{74.4} \\
\end{tabular}}}
   \vspace{-15pt}
\end{table}

%---------------------------------------------------------------
\subsection{Ablation Study on KITTI Dataset}
%---------------------------------------------------------------

\noindent\textbf{Multi-modality transformation flow.}\label{subsec:mmaug_exps}
We validate the effectiveness of \emph{multi-modality transformation flow} with different multi-modality augmentation techniques.
To obtain a clear comparison with SECOND~\cite{SECOND}, 
we conduct experiments on the vanilla MVX-Net without using the beneficial techniques discussed in Sec.~\ref{subsec:implement_detail}.
% Some points may not be visible in the pasted image after multi-modality cut and paste;
% for those points, we set their corresponding image features as zeros during multi-modality fusion.

The original MVX-Net~\cite{MVXNet} only uses global scaling in the augmentation, and global rotation and translation are thought as not applicable. 
\emph{Multi-modality transformation flow} enables random flipping, translation, and rotation to be applied for multi-modality detectors.
Table~\ref{tab:mmaug_add} verifies that random flipping, scaling, rotating, and translating point cloud are essential for both single-modality and multi-modality 3D detectors.
With these augmentations added sequentially, both the single-modality detector (SECOND~\cite{SECOND}) and the multi-modality detector (MVX-Net~\cite{MVXNet}) obtain significant improvement.

Notably, the results also show that \emph{multi-modality cut and paste} plays a critical role in
bridging the performance gap between the multi-modality detector and single-modality detector.
Both single-modality and multi-modality methods obtain large improvement with the help of cut and paste (moderate mAP from 60.7\% and 60.5\% to 68.1\% and 70.2\%, respectively).
%
%
% This indicates that existing models face severe overfitting problem in KITTI dataset and cut and paste could reduce this issue.
% Intuitively, the SECOND-based MVX-Net should surpass SECOND easily, which does not holds from Table~\ref{tab:mmaug_add}.
Without cut and paste, MVX-Net cannot even surpass SECOND (60.5\% vs. 60.7\%).
But with the help of \emph{MoCa}, MVX-Net surpasses its counterpart by a large margin (70.2\% vs. 68.1\%).

\begin{table}[t]
    \centering
    \caption{\small{Comparison of Faster R-CNN trained with different augmentations and different epochs. 
    `Baseline'-- not using cut and paste. 
    and `AutoAug'-- searched augmentation strategies~\cite{zoph2019learning}}}\label{tab:mmaug_epoch}
    \vspace{-9pt}
    \addtolength\tabcolsep{-0.3em}
    \scalebox{0.85}{\begin{tabular}{c|c|c|c|c}
        % Method & Epochs & $AP_{easy}$ & $AP_{mod.}$ & $AP_{hard}$ \\
        \multirow{2}{*}{Method} & \multirow{2}{*}{Epochs} & \multicolumn{3}{c}{2D mAP (\%)} \\ \cline{3-5}
        && Easy & Mod. & Hard \\ %\hline
        \hline\multirow{2}{*}{Baseline}
        & 20 & 86.3 & 75.2 & 71.8 \\
        & 36 & 86.6 & 73.3 & 67.4 \\
        \hline
        \multirow{2}{*}{AutoAug~\cite{zoph2019learning}} & 20 & 86.4  & 78.5 & 73.7 \\
        & 36 & 84.9&	78.2&72.1 \\
        \hline \multirow{2}{*}{\centering{MoCa (ours)}}
        & 20 & 86.3 & 79.3 & \textbf{74.4} \\
        & 36 & \textbf{86.6} & \textbf{79.5} & 74.1 \\
    \end{tabular}}
    \vspace{-18pt}
\end{table}

\noindent\textbf{Multi-modality cut and paste (MoCa).}
We evaluate the effectiveness of components in MoCa.
We use the instance masks in KINS dataset~\cite{qi2019amodal} to build the multi-modality GT database.
Note that we \emph{only use the objects in the training split to formulate the GT database in all experiments}.
No information from validation/test split is used in training.
% We empirically find that making the frame to have 6, 6, and 12 objects of pedestrian, cyclist, and car in training, respectively, leads to the best performance.
We empirically find that having 6 pedestrians, 6 cyclists, and 12 cars in a frame for training yields the best performance.
Table~\ref{tab:mmaug_thr} shows that collision test improves the detector's performance of both multi-modality 3D detector and 2D detector, and different thresholds lead to different APs under different conditions.
The proposed mixed IoF thresholds yield the best performance and is adopted in other experiments.

We also evaluate the effectiveness of MoCa for 2D detectors.
We train Faster R-CNN~\cite{ren2015faster} with FPN~\cite{lin2017_fpn} following the standard setting~\cite{mmdetection}
except that we use 20 or 36 epochs and different augmentation strategies.
Cut and paste significantly improves the results, in comparison with those not using cut and paste
and those using searched augmentation strategies~\cite{zoph2019learning} (Table~\ref{tab:mmaug_epoch}).
As the training schedule becomes longer, cut and paste maintains its high performance,
while the performances of baseline and AutoAug~\cite{zoph2019learning} degrade.
The results of sustaining a longer training suggest the effectiveness of MoCa in reducing overfitting.

\begin{table}[t]
	\small{\caption{
        Comparison of different training strategies.
        `Pre-train' indicates which pre-trained model is used.
        `COCO' means the model is pre-trained on COCO dataset, and `+ KITTI' means the model is further pre-trained on KITTI 2D dataset.
        `Freeze' indicates whether ResNet-50~\cite{He_2016} in the image branch is frozen.
        The average results and standard deviation over 5 runs are reported.
        The best results in each setting are bolded % and the best results across all the settings are marked bolded
	}\label{tab:cocktail_optimization}}
	\vspace{-21pt}
    \begin{center}
        \addtolength\tabcolsep{-0.3em}
		\scalebox{0.85}{\begin{tabular}{c|c|c|c|c|c}
            \multirow{2}{*}{Pre-train} &\multirow{2}{*}{Freeze}& \multirow{2}{*}{Optimizer} & \multicolumn{3}{c}{3D mAP (\%)}\\ \cline{4-6}
            &&&Easy & Mod. & Hard \\ \hline
            \multirow{3}{*}{COCO} &\multirow{3}{*}{$\times$}&SGD&$80.2\pm1.0$&$69.9\pm0.8$&$66.1\pm1.1$\\
            &&ADAM&$80.9\pm0.7$&$70.2\pm0.7$&$66.3\pm1.2$\\
            &&Hybrid&$\mathbf{81.1\pm0.7}$&$\mathbf{71.0\pm0.6}$&$\mathbf{67.2\pm0.6}$ \\\hline
            \multirow{6}{*}{\tabincell{c}{COCO\\+ \\KITTI}} &\multirow{3}{*}{$\times$}& SGD &$80.5\pm 0.8$&$70.3\pm 0.2$&$67.0\pm0.4$\\
            &&ADAM&$80.6\pm 0.3$&$69.6\pm 0.3$&$65.7\pm0.6$\\
            &&Hybrid&$\mathbf{81.1\pm 0.5}$&$\mathbf{70.6\pm 0.7}$&$\mathbf{67.0\pm1.0}$\\ \cline{2-6}
            &\multirow{3}{*}{\checkmark}&SGD&$80.3\pm 0.6$&$70.3\pm 0.4$&$66.5\pm0.8$\\
            &&ADAM&$80.9\pm 0.6$&$70.7\pm 0.5$&$67.0\pm0.3$\\
            &&Hybrid&$\mathbf{81.8\pm 0.4}$&$\mathbf{71.2\pm 0.5}$&$\mathbf{67.8\pm0.7}$\\
		\end{tabular}}
	\end{center}
	\vspace{-21pt}
\end{table}

\noindent\textbf{Retaining the original optimizers.}\label{subsec:cocktail_optim_exps}
As discussed in Sec.~\ref{subsec:implement_detail}, the choice of optimizer for the image feature extractor matters.
Here, we compare the following variants: (1) \textit{SGD}, the SGD optimizer is used for the image-branch pre-training as well as for both the image and point cloud branches during joint training; (2) \textit{ADAM}, the SGD optimizer is used for the image-branch pre-training but the ADAM optimizer is used for both the image and point cloud branches during joint training; (3) \textit{Hybrid}, we retain the original optimizer of each modality branch, as presented in Sec.~\ref{subsec:implement_detail}.
The optimal learning rate and other hyper-parameters for all the aforementioned variants are found using a grid search to ensure fair comparisons.
Furthermore, different pre-trained image feature extractors are tested to ensure completeness. More details on pre-training can be found in the appendix. 
As shown in Table~\ref{tab:cocktail_optimization}, retaining the original optimizers is beneficial.
Our results here are purely empirical but they reveal the interesting tendency of using different optimizers in different modality branches. This phenomenon worths further exploration especially from the perspective of optimization routes~\cite{SWATS, ZhouF0XHE20}.

\noindent\textbf{Step-by-step results.}
We evaluate the beneficial components step by step in Table~\ref{tab:fusion_design}.
We freezes image branch pre-trained on COCO and KITTI dataset when retaining the original optimizers (setting of last row in Table~\ref{tab:cocktail_optimization}).
MVX-Net enhanced by our methods significantly surpasses its previous version (60.5\%) by \textbf{9.5\%}, \textbf{11.3\%}, and \textbf{10.4\%} in mAP of easy, moderate, and hard conditions, respectively.
The enhanced MVX-Net also surpasses its single-modality counterpart, SECOND~\cite{SECOND}, by \textbf{3.1\%}, \textbf{3.7\%}, and \textbf{6.1\%} in mAP of easy, moderate, and hard conditions, respectively.

\begin{table}[t]
	\caption{
        \small{Ablation study of each component on KITTI validation dataset. Modifications are added sequentially}
	}\label{tab:fusion_design}
	\vspace{-18pt}
    \begin{center}
        \addtolength\tabcolsep{-0.3em}
		\scalebox{0.85}{\begin{tabular}{c|c|c|c}
        \multirow{2}{*}{Method} & \multicolumn{3}{c}{3D mAP (\%)} \\ \cline{2-4}
        & Easy & Mod. & Hard \\ \hline
            SECOND~\cite{SECOND} &78.3&68.1 & 63.4 \\ \hline
            + image branch~\cite{MVXNet}&71.9& 60.5&59.1 \\
            + MoCa&80.9&70.2&65.2 \\
            + retain original optimizers &81.2&71.0&65.7 \\
            + aligned pyramid fusion&\textbf{81.4}&\textbf{71.8}&\textbf{69.5} \\
		\end{tabular}}
	\end{center}
	\vspace{-21pt}
\end{table}

\begin{table}[t]
	\caption{
        \small{Ablation study on nuScenes validation set. Modifications are added sequentially}
	}\label{tab:mvx_nus_ablation}
	\vspace{-18pt}
    \begin{center}
        \addtolength\tabcolsep{-0.3em}
		\scalebox{0.85}{\begin{tabular}{c|c|c}
            Method & NDS (\%) &mAP (\%)\\ \hline
            PointPillars~\cite{PointPillars} + FPN~\cite{lin2017_fpn} &53.4&40.1 \\ \hline
            + image branch~\cite{MVXNet}&54.6&41.8 \\
            + MoCa&55.0&43.0	 \\
            + retain original optimizers &57.7&47.3	\\
            + aligned pyramid fusion &\textbf{58.1}&\textbf{47.9} \\
		\end{tabular}}
	\end{center}
	\vspace{-24pt}
\end{table}

% \begin{table}[t]
% 	\caption{
%         \small{Comparison of image branches pre-trained by different detectors}
% 	}\label{tab:mvx_nus_pretrain}
% 	\vspace{-18pt}
%     \begin{center}
%         \addtolength\tabcolsep{-0.3em}
% 		\scalebox{0.87}{\begin{tabular}{c|c|c|c|c}
%             Detector &Faster R-CNN&Mask R-CNN&Cascade& HTC \\ \hline
%             NDS (\%)& 57.4& 57.6&57.9&\textbf{58.1}\\
% 		\end{tabular}}
% 	\end{center}
% 	\vspace{-27pt}
% \end{table}

\begin{figure}[t]
    \includegraphics[width=0.48\textwidth]{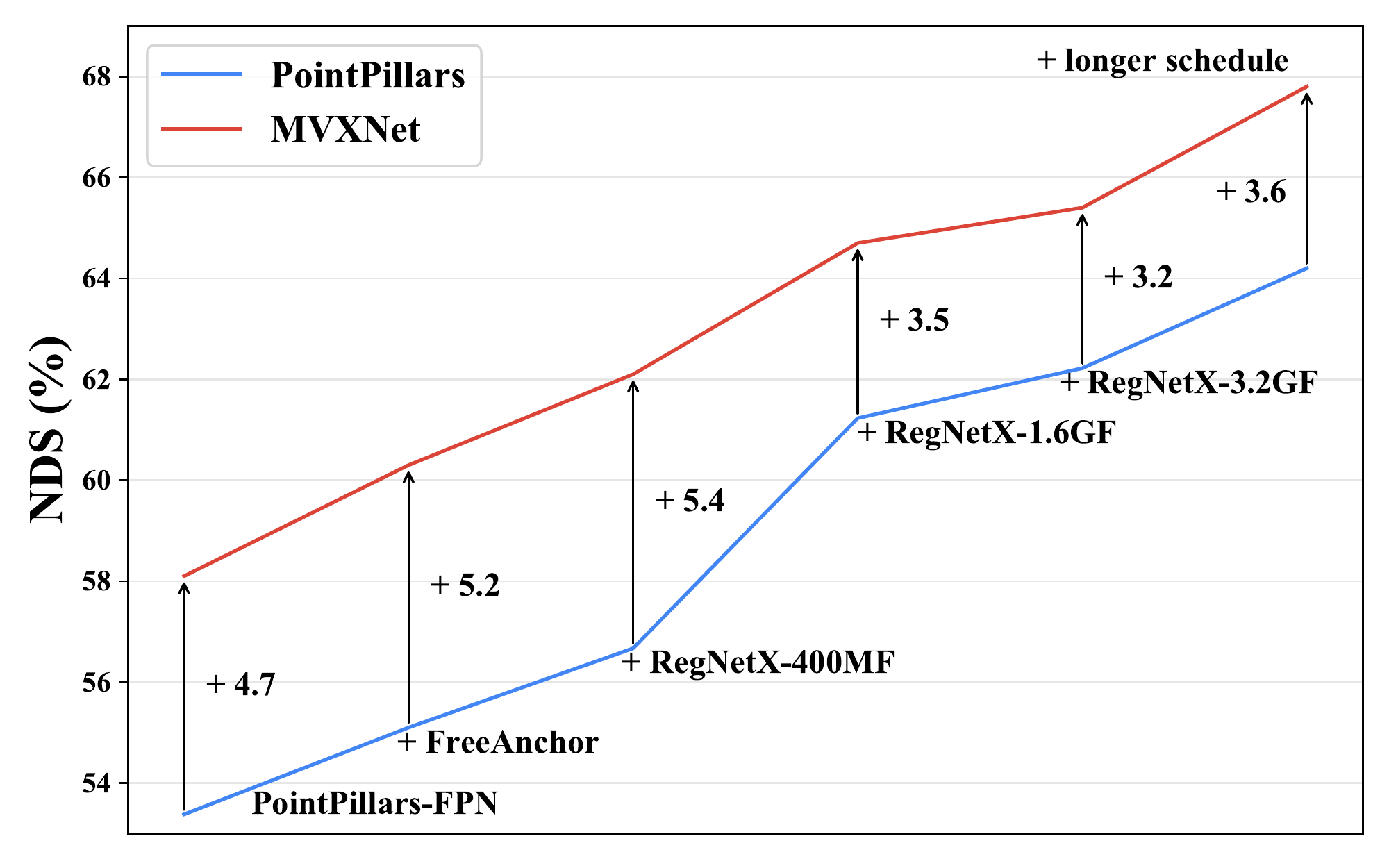}
    \vspace{-24pt}
    \caption{\small{Improvement by MVX-Net on different baselines.}
    }\label{fig:mvx_pp_compare}
	\vspace{-15pt}
\end{figure}

%---------------------------------------------------------------
\subsection{Ablation Study on nuScenes Dataset}
%---------------------------------------------------------------
We further validate MoCa and the enhanced MVX-Net on the more challenging nuScenes dataset.
Since existing results in the literature are mainly based on PointPillars~\cite{PointPillars}, we also adopt PointPillars as the point cloud branch in MVX-Net~\cite{MVXNet}.
We reimplement PointPillars~\cite{PointPillars} but supplement it with FPN~\cite{lin2017_fpn}.
PointPillars~\cite{PointPillars} with FPN~\cite{lin2017_fpn} achieves 53.4\% in the NDS score, higher than that reported by the dataset provider (44.2\% ~\cite{nuScenes}).
We empirically find the image branch pre-trained from HTC~\cite{Chen_2019_CVPR} performs better; thus, we use it in our experiments. More training details are in the appendix.

As shown in Table~\ref{tab:mvx_nus_ablation}, MoCa and the explored practices are all beneficial on nuScenes dataset. The proposed changes allow MVX-Net to surpass PointPillars by a large margin (7.8\% mAP and 4.7\% NDS).
%Out of our expectation, hybrid optimization exhibits more improvements on nuScenes dataset (3.9 mAP and 2.5\% NDS).

% \noindent\textbf{Pre-training.}
% We also observe different effectiveness of the image branch in our experiments when it is pre-trained in Faster R-CNN~\cite{ren2015faster}, Mask R-CNN~\cite{mask_rcnn}, Cascade Mask R-CNN~\cite{cascade_rcnn}, and HTC~\cite{Chen_2019_CVPR} (Table~\ref{tab:mvx_nus_pretrain}).
% Using similar backbones and necks, the image branch from HTC pre-trained on nuImages dataset~\cite{nuScenes} shows a gain of 0.7\% NDS against that from Faster R-CNN.
% Thus, we adopt the image branch from HTC in other experiments on nuScenes dataset.
% More details are provided in the appendix.

\noindent\textbf{Stronger PointPillars baselines.}
We further verify our methods over stronger baselines.
The original baseline is PointPillars with FPN in the neck. We replace the head with FreeAnchor head~\cite{FreeAnchor} to predict 3D boxes.
We further replace the backbone of PointPillars with RegNetX~\cite{regnet}, including RegNetX-400MF, RegNetX-1.6GF, and RegNetX-3.2GF.
As shown in Fig.~\ref{fig:mvx_pp_compare}, despite the increasing performance of PointPillars due to enhancements applied to its backbone, neck, head, and longer training schedule, the enhanced MVX-Net still improves the baseline consistently.
The results suggest the effectiveness and generalizability of our methods.

% !TEX root = ../main.tex

\section{Conclusion}
This paper investigates and discusses the pitfalls of applying data augmentations to multi-modality 3D object detection.
We contribute a pipeline, \emph{multi-modality transformation flow}, to ensure consistency during augmentations and to enable a richer set of augmentation strategies.
Based on that, we validate the effectiveness of different augmentations that are absent in previous works and further present \emph{multi-modality cut and paste (MoCa)} to bridge the gap of augmentations between multi-modality and single-modality 3D detectors.
Under different strong baselines, our method improves the performance consistently and achieves new state-of-the-art performance on nuScenes dataset.
%Our method also exhibits higher generalizability on the hard cases in both KITTI and nuScenes benchmarks.

\appendix
\begin{appendices}
% !TEX root = ../main.tex
\section{Implementation Details}
\noindent\textbf{Projection from LiDAR to image.}
For feature fusion, the point cloud is first transformed from LiDAR coordinates $P_{lidar}$ to camera coordinates and then projected to image pixel coordinates $P_{img}$.

On KITTI dataset~\cite{Geiger2012CVPR}, the projection is calculated as follows:
\begin{equation}
    P_{img} = P_{rect}^{0}R_{rect}^{0}T_{cam \leftarrow lidar}P_{lidar},
\end{equation}
where $T_{cam \leftarrow lidar}$ is the transformation matrix from LiDAR coordinates to
camera coordinates, $R_{rect}^{0}$ is the rectifying rotation matrix of the left camera, 
and $P_{rect}^{0}$ is the calibration matrix of the left camera.

On nuScenes dataset~\cite{nuScenes}, 
the points in LiDAR coordinates are first transformed to the ego car's global coordinates since the LiDAR (20Hz) and camera (12Hz) work at different frequencies.
Therefore, the transformation is calculated as follows:
\begin{equation}
    P_{img} = T_{cam \leftarrow ego}T_{ego_{c} \leftarrow ego_{l}}T_{ego \leftarrow lidar}P_{lidar},
\end{equation}
where $T_{ego \leftarrow lidar}$ is the transformation matrix from LiDAR to the ego pose at timestamp $t_l$ when the LiDAR frame is recorded,
$T_{ego_{c} \leftarrow ego_{l}}$ is the transformation matrix from ego pose at timestamp $t_l$ to the ego pose at timestamp $t_c$ when the camera frame is captured,
and $T_{cam \leftarrow ego}$ is the transformation matrix from ego pose at timestamp $t_c$ to the camera.

\noindent\textbf{Multi-modality cut and paste.}
Multi-modality cut and paste (MoCa) needs ground-truth instance masks to paste image patches of objects, as in~\cite{Dwibedi_2017_ICCV, Dvornik_2018_ECCV, instaboost}.
For KITTI dataset, we use the instance masks in KINS dataset~\cite{qi2019amodal} to build the multi-modality GT database.
For nuScenes dataset, we use pseudo instance masks predicted by an instance segmentation model.
Specifically, we first train an HTC~\cite{Chen_2019_CVPR} with ResNeXt 32$\times$4d backbone on COCO dataset with 3$\times$ schedule~\cite{mmdetection}.
Then we fine-tune the HTC on nuImages dataset by 20 epochs and use that model to predict instance masks of images in nuScenes dataset.

\noindent\textbf{Training details.}
On KITTI dataset, both SECOND~\cite{SECOND} and MVX-Net~\cite{MVXNet} are trained by 80 epochs with a batch size of 16.
We adopt a half-period cosine schedule~\cite{SGDR} for learning rate decaying and use a linear warm-up strategy in the first 1K iterations.
The initial learning rate is 0.003 for all the 3D detectors that use ADAM~\cite{ADAM} optimizer.
When retaining the original optimizers (Sec. 3.3) for MVX-Net, we train the image branch using SGD optimizer with momentum and the initial learning rate is 0.05.
The hyperparameters for point cloud branch remain the same as those for SECOND.

For nuScenes dataset~\cite{nuScenes},
both PointPillars~\cite{PointPillars} and MVX-Net~\cite{MVXNet} are trained by 20 epochs with batch sizes of 32 and 16, respectively.
Notably, this is different from the official implementation~\cite{nuScenes}, where PointPillars~\cite{PointPillars} is trained by 125 epochs, which costs much time.
We adopt step learning rate decaying schedule following the practice in mmdetection~\cite{mmdetection}, \ie, the learning rate is decayed by 0.1 after the 16th and 19th epoch, respectively.
The initial learning rate is 0.001 for all the 3D detectors using ADAM~\cite{ADAM} optimizer.
When training MVX-Net, we also retains the original optimizers, where the image branch is trained by SGD optimizer with momentum and the initial learning rate is 0.0012.
The hyperparameters for point cloud branch remain the same as those for PointPillars.

Because one training sample for MVX-Net contains six images from multiple views and one LiDAR point cloud frame, each GPU can only contain one training sample during each iteration.
This degrades the performance siginificantly because the batch size in one GPU is so small that the inaccurate statistics in Batch Normalization (BN)~\cite{BN} affects the training process.
Therefore, we use Syncronized Batch Normalization (SyncBN)~\cite{liu2018_panet} to solve this issue.
We report all the results of our methods using SyncBN in Table 2 and 8 on nuScenes dataset of the main paper.
We do not use SyncBN for models on KITTI dataset because the batch size is 2 in each GPU as there is only one image and one LiDAR point cloud frame in one training sample.

\vspace{-3pt}
\section{Experiments}
\begin{table}[t]
	\caption{
        \small{Comparison of image branches pre-trained by different detectors}
	}\label{tab:mvx_nus_pretrain}
	\vspace{-18pt}
    \begin{center}
        \addtolength\tabcolsep{-0.3em}
		\scalebox{0.87}{\begin{tabular}{c|c|c|c|c}
            Detector &Faster R-CNN&Mask R-CNN&Cascade& HTC \\ \hline
            NDS (\%)& 57.4& 57.6&57.9&\textbf{58.1}\\
		\end{tabular}}
	\end{center}
	\vspace{-27pt}
\end{table}

\begin{table*}[!h]
	\caption{
        \small{Step-by-step results of stronger baselines on nuScenes validation set.
        `Con. Veh.', `Ped.', and `T.C.' are the abbreviations of construction vehicle, pedestrian, and traffic cone, respectively.
        `FA' means FreeAnchor~\cite{FreeAnchor} and `3$\times$' means longer training schedule.
        NDS score, mAP, and APs of each categories are reported. The best results are bolded}
	}\label{tab:strong_baselines}
	\vspace{-18pt}
    \begin{center}
        \addtolength\tabcolsep{-0.3em}
		\scalebox{0.87}{\begin{tabular}{c|c|c|c|c|c|c|c|c|c|c|c|c|c}
            Method &Modality& NDS (\%) &mAP (\%) &Car& Truck & Bus & Trailer & Con. Veh. &Ped. & Motor & Bicycle  & T.C. & Barrier \\ \hline
            PointPillars~\cite{PointPillars} + FPN~\cite{lin2017_fpn}&\multirow{6}{*}{L}&53.4&40.1&80.6&35.9&43.5&29.2&5.4&71.9&34.9&11.8&35&52.6 \\
            + FA~\cite{FreeAnchor}&&55.1&43.7&81.5&40.0&50.0&29.4&9.2&74.3&44.5&16.5&39.6&52.4 \\
            + FA + RegNetX-400MF\cite{regnet}  &&56.7&45.5&82.0&41.9&50.7&32.3&11.0&75.4&50.1&19.1&44.1&48.8 \\
            + FA + RegNetX-1.6GF &&61.2&51.4&83.2&48.2&60.5&30.4&16.6&78.1&59.4&25.9&49.1&62.3	 \\
            + FA + RegNetX-3.2GF &&62.2&52.1&83.6&51.1&62.3&36.0&17.3&78.2&56.1&24.7&50.0&62.0 \\
            + FA + RegNetX-3.2GF + 3$\times$~ &&64.2&56.9&85.5&54.9&66.8&35.4&22.2&81.2&62.4&35.6&59.2&65.4	 \\ \hline
            MoCa &\multirow{6}{*}{L + I}&58.1&47.9&82.4&41.5&49.6&28.6&9.1&79.1&50.3&27.2&49&61.9 \\
            + FA~\cite{FreeAnchor}    &&60.3&52.9&83.6&48.5&56.4&31.4&10.8&81.6&61.0&35.7&58.1&61.7 \\
            + FA + RegNetX-400MF\cite{regnet}  &&62.1&55.2&84.2&51.7&63.6&34.2&18.0&82.0&61.8&32.9&59.4&64.0 \\
            + FA + RegNetX-1.6GF &&64.7&	58.4& 85.2&55.5&64.5&35.0&20.8&84.5&68.0&42.8&62.3&65.1 \\
            + FA + RegNetX-3.2GF &&65.4&59.2&85.7&56.8&66.2&35.8&21.7&84.4&67.4&44.2&62.8&66.3 \\
            + FA + RegNetX-3.2GF + 3$\times$~ &&\textbf{67.8}&\textbf{62.5}&\textbf{87.0}&\textbf{60.8}&\textbf{68.2}&\textbf{39.7}&\textbf{23.9}&\textbf{85.8}&\textbf{70.2}&\textbf{51.6}&\textbf{68.1}&\textbf{69.5}	 \\
		\end{tabular}}
	\end{center}
	\vspace{-18pt}
\end{table*}

\begin{table*}[!h]
	\caption{
        \small{Performance of top-3 submissions to the 3rd nuScenes Detection Challenge on test set.
        `Con. Veh.', `Ped.', and `T.C.' are the abbreviations of construction vehicle, pedestrian, and traffic cone, respectively.
        `I', `L', and `R' means whether data from image, LiDAR, and RADAR is used, respectively.
        NDS score, PKL~\cite{PKL}, mAP, and APs of each categories are reported}
	}\label{tab:challenge}
	\vspace{-18pt}
    \begin{center}
        \addtolength\tabcolsep{-0.3em}
		\scalebox{0.87}{\begin{tabular}{c|c|c|c|c|c|c|c|c|c|c|c|c|c|c}
            Method  &Modality&PKL& NDS (\%) &mAP (\%) &Car& Truck & Bus & Trailer & Con. Veh. &Ped. & Motor & Bicycle  & T.C. & Barrier \\ \hline
            CenterPoint v2~\cite{CenterPoint, pointpainting}& I + L + R&0.581&71.4&67.1&87.0&57.3&69.3&60.4&28.8&90.4&71.3&49.0&86.8&71.0 \\
            PointAugmenting & I + L &0.595&71.1&66.8&87.5&57.3&65.2&60.7&28.0&87.9&74.3&50.9&83.6&72.6 \\
            Ours& I + L  &0.574&70.9&66.6&86.7&58.6&67.2&60.3&32.6&87.1&67.8&52.0&81.3&72.3
		\end{tabular}}
	\end{center}
	\vspace{-18pt}
\end{table*}

\noindent\textbf{Retaining the original optimizers.}
In Table 6 of the main paper, to verify the versatility of retaining the original optimizers and find a good training strategy,
we adopt three training strategies for pre-training image feature extractors and jointly training multi-modality detectors.
The first strategy trains a Faster R-CNN~\cite{ren2015faster} on COCO2017 dataset by the multi-scale $3\times$ schedule~\cite{mmdetection, wu2019detectron2} with ResNet-50~\cite{He_2016} pre-trained on ImageNet~\cite{ILSVRC15}.
Then we train Faster R-CNN on the subset of COCO training split that only contains three classes: people (for pedestrian in KITTI), bicycle (for cyclist in KITTI), and car, following Frustum-PointNet~\cite{fpointnet}.
The weights of ResNet-50 and FPN pre-trained in the Faster R-CNN are adopted in the image branch of the multi-modality detector for joint training.
The second strategy further fine-tunes the Faster R-CNN using \emph{MoCa} for 20 epochs with mixed IoF thresholds (Table 4 of the main paper) before joint training.
The third strategy uses a similar pre-training strategy as the second one but freezes the ResNet-50 backbone in joint training.
Table 6 shows that the synergy of the third strategy and retaining the original optimizers works best among these training and optimization strategies.

\noindent\textbf{Pre-training on nuScenes dataset.}
We observe different effectiveness of the image branch in our experiments when it is pre-trained in Faster R-CNN~\cite{ren2015faster}, Mask R-CNN~\cite{mask_rcnn}, Cascade Mask R-CNN~\cite{cascade_rcnn}, and HTC~\cite{Chen_2019_CVPR} (Table~\ref{tab:mvx_nus_pretrain}).
Using similar backbones and necks, the image branch from HTC pre-trained on nuImages dataset~\cite{nuScenes} shows a gain of 0.7\% NDS against that from Faster R-CNN.
Thus, we adopt the image branch from HTC in other experiments on nuScenes dataset.

The image branch from HTC pre-trained on nuImages dataset~\cite{nuScenes} shows a gain of 0.7\% NDS against that from Faster R-CNN as shown in Table~\ref{tab:mvx_nus_pretrain}.
Those pre-trained detectors are first trained by 20 epochs using the standard setting on COCO dataset and released by MMDetection~\cite{mmdetection}.
We further fine-tune those detectors on the nuImages dataset~\cite{nuScenes} by 20 epochs using similar hyperparameters as those for the COCO dataset.
When jointly training the multi-modality detectors, the weights in ResNet-50 backbone and FPN of those pre-trained detectors are adopted to initialize the image branch.

\noindent\textbf{Stronger PointPillars baselines.}
We adopt stronger PointPillars baselines in MVX-Net to verify our method's generalizability by enhancing the head, backbone, and training schedule of PointPillars.
For completeness, we also put the detailed step-by-step results of Figure 6 in Table~\ref{tab:strong_baselines}.
Notably, our PointPillars baseline already achieves very high performance on the validation set.
However, our methods consistently improve performance,
especially on challenging classes for LiDAR-based detectors, such as bicycles, motorcycles, traffic cones, and pedestrians.

\section{nuScenes Detection Challenge}

Based on the model that achieves the best result in  Table~\ref{tab:mvx_nus_val} (last row), we ensemble the other five models with test-time augmentation and submit the results
of test set to the 3rd nuScenes detection challenge.
The challenge introduces Planning KL-Divergence (PKL)~\cite{PKL}, a novel planning-based metric that measures the influence of predictions on the downstream tasks in autonomous driving.
The lower value of PKL means that the predictions are more similar to ground truth and are more suitable for planning.
We list the top-3 entries including our submission of the challenge in Table~\ref{tab:challenge}.
The results show that our submission achieves the best PKL results on the test set, which has the least deviation from the ground truth and is more favorable for autonomous driving.

\end{appendices}

%%%%%%%%%%%%%%%%%%%%%%%%%%%%%%%%%%%%%%%%%%%%%%%%%%%%%%%%%%%%%%%%%%%%%%%%%%%%%%%%%%%%%%%%%%%%%%%%%%%

{\small
	\bibliographystyle{ieee_fullname}
	\bibliography{sections/mainbib}
}

\end{document}